\documentclass{article}

\usepackage{microtype}
\usepackage{graphicx}
\usepackage{subcaption}
\usepackage{booktabs} 

\usepackage{hyperref}

\usepackage[preprint]{icml2026}

\usepackage{amsmath}
\usepackage{amssymb}
\usepackage{mathtools}
\usepackage{amsthm}

\begin{document}

\twocolumn[
\icmltitle{From independent patches to coordinated attention: Controlling information flow in vision transformers}

\begin{icmlauthorlist}
\icmlauthor{Kieran A. Murphy}{njit}
\end{icmlauthorlist}

\icmlaffiliation{njit}{Department of Computer Science, New Jersey Institute of Technology, Newark, NJ, USA 97102}

\icmlcorrespondingauthor{}{kieran.murphy@njit.edu}

\vskip 0.3in
]

\printAffiliationsAndNotice{}  

\begin{abstract}

We make the information transmitted by attention an explicit, measurable quantity in vision transformers.
By inserting variational information bottlenecks on all attention-mediated writes to the residual stream---without other architectural changes---we train models with an explicit information cost and obtain a controllable spectrum from independent patch processing to fully expressive global attention.
On ImageNet-100, we characterize how classification behavior and information routing evolve across this spectrum, and provide initial insights into how global visual representations emerge from local patch processing by analyzing the first attention heads that transmit information.
By biasing learning toward solutions with constrained internal communication, our approach yields models that are more tractable for mechanistic analysis and more amenable to control.

\end{abstract}

\section{Introduction}
\label{introduction}

Transformers have emerged as a dominant architectural paradigm across machine learning, distinguished by their ability to aggregate information globally across the constitutive elements of an input \cite{vaswani2017}.
In vision transformers \cite{dosovitskiy2021}, an image is decomposed into patches, and computation proceeds through repeated alternation between two modules: attention blocks, which enable information sharing across patches, and multilayer perceptron (MLP) blocks, which refine patch representations independently.
Because communication and local refinement occur in distinct computational steps, information flow becomes an explicit and manipulable axis of the architecture, rather than being inextricably entangled with processing as in prior paradigms.
This naturally raises a mechanistic question: when communication is restricted, what circuits of attention-mediated interaction emerge first~\cite{olah2020zoom,elhage2021mathematical}, and what do they reveal about how transformers build global understanding from local evidence?

Modern interpretability methods typically analyze models trained without any explicit pressure toward simplicity \cite{rudin2019stop,samek2021xai} or prescribed explanatory structure (e.g., prototype-based or concept-bottleneck constraints) \cite{chen2019looks,koh2020cbm}, often revealing internal mechanisms that are rich but difficult to disentangle or control.
A complementary approach is to bias learning itself toward solutions with constrained information flow \cite{alemiVIB2016}.
However, properly reasoning about the movement of information requires going beyond the deterministic, point-valued transformations that dominate standard architectures, which make it difficult to assign meaning to ``information'' in an operational sense \cite{saxe2019}.
One must instead introduce stochastic or probabilistic representations that allow information to be measured and restricted.

\begin{figure*}[ht]
\vskip 0.2in
\begin{center}
\centerline{\includegraphics[width=\textwidth]{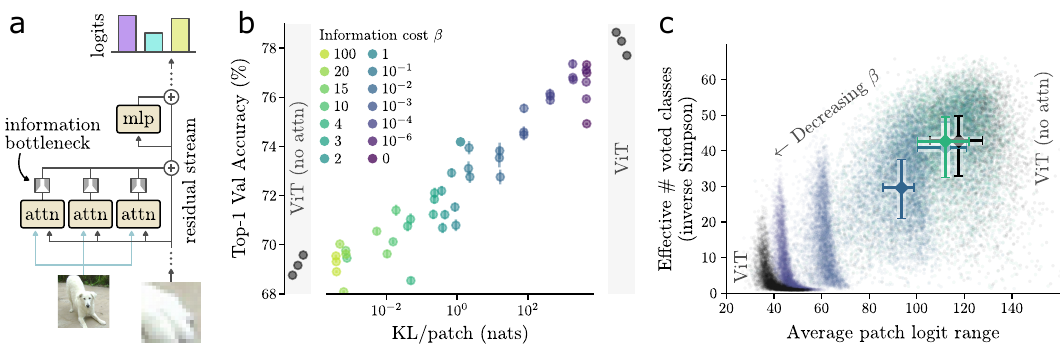}}
\caption{\textbf{Limiting information written by attention to the residual stream.}
\textbf{(a)} We install an information bottleneck (IB) after every attention head and before anything is written to a patch's residual stream.  
The sum total of information penalties is added to the original training loss, scaled by a parameter $\beta$ that induces a spectrum from independent patch voting (no attention-mediated communication) to free-flowing information as in an unmodified ViT.
\textbf{(b)} By installing IBs after each of the 36 attention heads in a ViT-tiny trained on a subset of Imagenet, we obtain a spectrum of models parameterized by the total amount of information written to the residual stream by attention heads. 
As the amount of information increases, so too does accuracy, smoothly covering the span between the ViT without attention (left) and the unmodified ViT (right).
Error bars are shown only for the stochastic IB models (standard deviation across 10 validation runs).
\textbf{(c)} The voting behavior of patches in a single image also smoothly interpolates between the two extremes as a function of $\beta$.
We measure the average range of patch logits (min to max) and the variety of top-assigned classes across patches, using the inverse Simpson index as an effective count of diversity.
Every dot represents an image from the dataset (colored by $\beta$ with the same mapping as in \textbf{b}), and the median with interquartile ranges is shown for the point clouds with significant overlap.
}
\label{fig:fig1}
\end{center}
\vskip -0.2in
\end{figure*}

Such information-theoretic modifications to bottleneck the flow of information can be applied either post-hoc or during training.
Post-hoc bottleneck methods \cite{Schulz2020IBA,jiang2020IBAEMNLP,Hong2025CoIBA} enable the study of pretrained models, but necessarily probe them under perturbations that differ from the conditions under which they were optimized, raising concerns about faithfulness.
In contrast, training-time information restriction directly shapes the solutions that optimization discovers \cite{alemiVIB2016,achillesoatto2018}, yielding models whose internal computations are intrinsically biased toward limited communication.
In this work, we pursue this second approach, using vision transformers as a natural setting in which to study restricted information routing in a clean and controlled way.

Our goal is a minimal modification to the transformer architecture that allows attention-mediated communication to be smoothly controlled, producing a spectrum of models that interpolates between fully expressive transformer computation and attention-less independent patch processing.
We achieve this by inserting variational information bottlenecks on the writes to the residual stream performed by attention blocks, without otherwise altering the attention mechanism or transformer structure.
This yields a family of models in which communication capacity becomes a tunable parameter.

At the zero-communication limit, the architecture reduces to independent per-patch processing followed by permutation-invariant aggregation, closely related to the DeepSets formulation~\cite{deepsets}.
This provides a well-defined and interpretable endpoint.
As communication capacity increases, global visual representations emerge progressively from local patch computations.
By analyzing performance, patch-level information updates, and the first attention heads that become active, we obtain a mechanistic picture of how transformers begin to integrate information across space under constrained bandwidth.

\paragraph{Contributions.}
Concretely, we:
(i) introduce a controllable family of vision transformers with variational bottlenecks on every attention-mediated write to the residual stream;
(ii) characterize the tradeoff between accuracy and routed information across a spectrum from independent patch processing to standard ViTs on ImageNet-100;
and (iii) provide a fine-grained analysis of the first attention heads that transmit information, illustrating how global computation emerges from restricted local processing.
Overall, our results suggest that training-time information restriction yields models that are intrinsically more interpretable and more amenable to analysis and control.

\section{Related work}
\label{related_work}

\paragraph{Mechanistic interpretability and transformer circuits.}
The aim of this work aligns with mechanistic interpretability: to understand and ultimately control the internal information processing of transformer models.
Two complementary directions have emerged in this literature.
The first traces the flow of information between sequence elements as mediated by attention, identifying structured computational subgraphs or ``circuits'' within trained networks~\cite{olah2020zoom,elhage2021mathematical}.
The second seeks to decompose high-dimensional representation spaces into more interpretable features or concepts, for example through sparse dictionary learning or superposition-based analyses \cite{bricken2023monosemanticity,huben2024sparse}.
Because full-scale models are challenging to analyze directly, many mechanistic insights have been developed in simplified settings, including small or partially linearized transformers \cite{elhage2021mathematical} and narrowly defined tasks \cite{akyurek2023what,nanda2023grok,wang2023ioi,dutta2024how}.
Our work contributes a complementary form of simplification: rather than restricting model size or task complexity, we restrict internal communication capacity during training, yielding a controllable spectrum of transformer computations.

\paragraph{Mechanistic interpretability beyond language.}
Most mechanistic interpretability research has focused on autoregressive language models, where attention patterns and discrete token structure provide a natural substrate for circuit analysis.
Extending these ideas to vision transformers is less mature: sequence elements correspond to continuous image patches rather than symbols, attention is typically dense rather than masked, and the relevant computations are harder to localize in semantic units. 
Nevertheless, several lines of work have begun to probe the internal organization of ViTs, including representation-level studies of how transformer features evolve with depth and differ from CNNs~\cite{raghu2021vision,caron2021emerging,park2023what,colors2classes}, and feature-decomposition approaches such as sparse autoencoders applied to ViT activations~\cite{stevens2025saevit}.
These differ from attribution-based explainability methods\cite{chefer2021deeptaylor,covert2023learning,lrp_transformers_2024} in that they aim to characterize internal representations, but they still analyze models after training rather than shaping information flow during learning. 
In contrast, we study how global visual representations emerge from local patch processing when attention-mediated communication is explicitly restricted during optimization.

\paragraph{Information bottleneck methods for interpretability.}
The methodology of this work is closely related to the information bottleneck principle \cite{tishbyIB2000,alemiVIB2016}, which formalizes learning as a tradeoff between predictive sufficiency and compression.
While the information bottleneck has been widely used as a regularizer or robustness tool \cite{hu2024survey}, a distinct line of work uses bottlenecks to reveal which information is most important to a model's computation.
Several approaches apply information bottlenecks post-hoc to pretrained networks, including the Information Bottleneck Attribution method (IBA)~\cite{Schulz2020IBA} and subsequent extensions~\cite{jiang2020IBAEMNLP,Hong2025CoIBA}.
Other approaches incorporate bottlenecks directly during training, enabling analysis in the same regime in which the model was optimized.
For example, distributed information bottleneck methods have been used to quantify feature importance by restricting communication about input dimensions in tabular settings~\cite{dib_iclr,dib_pnas}.
Our work extends this training-time bottleneck perspective to the internal communication pathways of transformers by placing variational bottlenecks on attention-mediated residual writes.

\paragraph{Communication restriction and structured simplification in transformers.}
A closely related set of ideas seeks to simplify transformer computation by limiting or pruning attention and token interactions.
Prior work has explored attention head pruning and sparsification \cite{michel2019pruning,voita2019analyzing}, as well as token pruning and adaptive computation in vision transformers \cite{pan2021ia,rao2021dynamicvit,liang2022evit}.
These approaches primarily aim to improve efficiency, whereas our goal is interpretability: by making communication costly during training, we obtain a spectrum from independent patch processing to fully expressive attention, and can analyze which attention heads and latent dimensions first become active as capacity increases.

\begin{figure}[ht]
\vskip 0.2in
\begin{center}
\centerline{\includegraphics[width=\columnwidth]{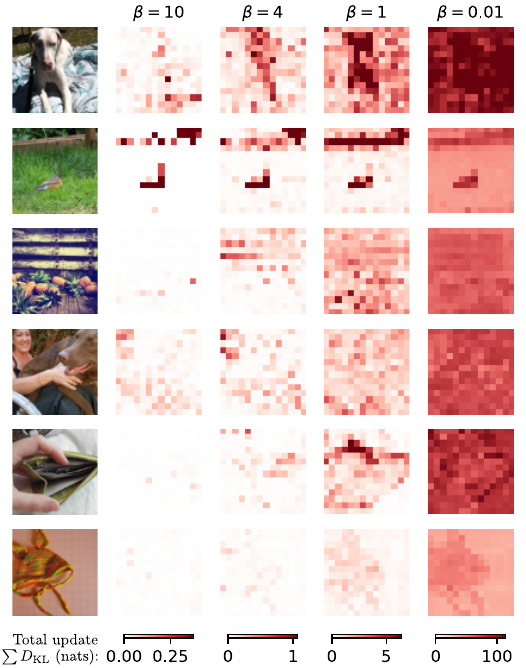}}
\caption{\textbf{KL allocation across patches.}
For a random sample of validation images that ViT correctly classified, we show the total KL per patch (summed across all attention heads in the model) for a selection of models nearest to the Pareto front in Fig.~\ref{fig:fig1}.
Note that the colormaps have the same range for each model (column).}
\label{fig:fig2}
\end{center}
\vskip -0.2in
\end{figure}

\section{Method}
\label{method}

We begin with a brief review of the vision transformer architecture, then introduce our training-time information restriction mechanism, and finally describe how we characterize the resulting spectrum of information routing behaviors.

\subsection{Vision transformers as residual stream communication}
We adopt the residual stream framing of transformer computation developed in \citet{elhage2021mathematical}.
A vision transformer (ViT) processes an image as a sequence of patch embeddings,
$\mathcal{S} = \{x_i\}_{i=1}^N$, where each patch maintains its own residual stream representation throughout the network.
Each transformer block alternates between two update mechanisms:
(i) an attention block, which computes context-dependent interactions across patches, and
(ii) an MLP block, which refines each patch representation independently.

Crucially, only attention enables information exchange between sequence elements (Fig.~\ref{fig:fig1}a).
MLP blocks operate pointwise on each residual stream and introduce no new cross-patch information.
Thus, attention-mediated writes to the residual stream constitute the natural pathway through which global visual representations emerge from local patch features.
Our approach directly monitors and restricts this communication channel.

\subsection{Bottlenecking attention-mediated residual writes}
Within an attention block, there are multiple possible intervention points at which one could restrict information flow.
We choose the final attention update immediately before it is written back into the residual stream.
This placement leaves the attention computation itself intact, while controlling the amount of information that can be transmitted through the update.
To obtain head-resolved control over communication, we instantiate a separate variational bottleneck on every attention head in every layer (36 bottlenecks total in ViT-Tiny), so that each head constitutes an independently regularized information channel.

Concretely, let $\Delta_i^\ell \in \mathbb{R}^d$ denote the attention update for patch $i$ at layer $\ell$ prior to the output projection.
In a standard transformer block, this update is added deterministically to the residual stream.
Instead, we route $\Delta_i^\ell$ through a variational information bottleneck:
\begin{equation}
\Delta_i^\ell \;\rightarrow\; z_i^\ell \sim q_\phi(z \mid \Delta_i^\ell)
\;\rightarrow\; \hat{\Delta}_i^\ell = g_\theta(z_i^\ell),
\end{equation}
where $q_\phi$ is an encoder network producing a stochastic latent representation, and $g_\theta$ is a decoder mapping back to the original update space.
The decoded update $\hat{\Delta}_i^\ell$ is then projected and written to the residual stream as usual.
By intervening only at the attention-mediated residual write---the sole mechanism by which patches exchange information---we obtain a deliberately minimal modification that isolates inter-patch communication without otherwise altering ViT computation.
This targeted design yields a controlled and interpretable axis of variation in global information routing.

\subsection{Training objective and information restriction}
Our bottlenecks are trained using the variational information bottleneck (VIB) principle~\cite{tishbyIB2000,alemiVIB2016}.
Each latent variable is modeled as a diagonal Gaussian,
\begin{equation}
q_\phi(z \mid \Delta) = \mathcal{N}\big(\mu_\phi(\Delta), \mathrm{diag}(\sigma_\phi^2(\Delta))\big),
\end{equation}
and is regularized toward an isotropic unit Gaussian prior $p(z)=\mathcal{N}(0,I)$.
The resulting KL divergence upper bounds the mutual information between $\Delta$ and $z$, yielding a tractable measure of transmitted information.

We train the model with the objective
\begin{equation}
\mathcal{L}=\mathcal{L}_{\mathrm{cls}}+\beta \sum_{\ell,i} 
\mathbb{E}_{q_\phi(z_i^\ell \mid \Delta_i^\ell)}
\big[ D_{\mathrm{KL}}\big(q_\phi(z_i^\ell \mid \Delta_i^\ell)\,\|\,p(z)\big) \big],
\label{eq:vib_objective}
\end{equation}
where $\mathcal{L}_{\mathrm{cls}}$ is the standard cross-entropy loss and $\beta$ controls the strength of the information restriction.
Varying $\beta$ therefore provides a single, interpretable control knob that continuously tunes the capacity of the network’s communication channel, from independent patch processing to the fully expressive ViT.

Although applying bottlenecks simultaneously across many layers departs from the classical single-bottleneck IB formulation, it provides a principled restriction on the total \emph{variety} of attention-mediated updates that can be transmitted through the network.
Moreover, the resulting IB latent spaces admit rich information-theoretic and mechanistic analyses.

\paragraph{Comparing probabilistic head updates via normalized mutual information.}
To compare the information content of two attention-head bottlenecks, we treat each head as inducing a soft clustering assignment over patches~\cite{nmi2024}.
For a given head, let $X$ denote the discrete patch identity (indexed over all patches in the validation set) and let $U$ denote the corresponding bottleneck latent variable.
Each head therefore defines a stochastic channel $p(u|x)$ through its variational posterior.

We estimate the mutual information between patch identities and latent messages using a Monte Carlo approximation based on the aggregated posterior over a dataset of size $L$:
\begin{equation}
    I(X;U) = \mathbb{E}_{x \sim p(x)} \mathbb{E}_{u \sim p(u|x)} \left [ \log \frac{p(u|x)}{\frac{1}{L}\sum_i^L p(u|x_i)} \right ],
\end{equation}
where the denominator provides an empirical estimate of the marginal $p(u)$.

\begin{figure}
\begin{center}
\centerline{\includegraphics[width=\columnwidth]{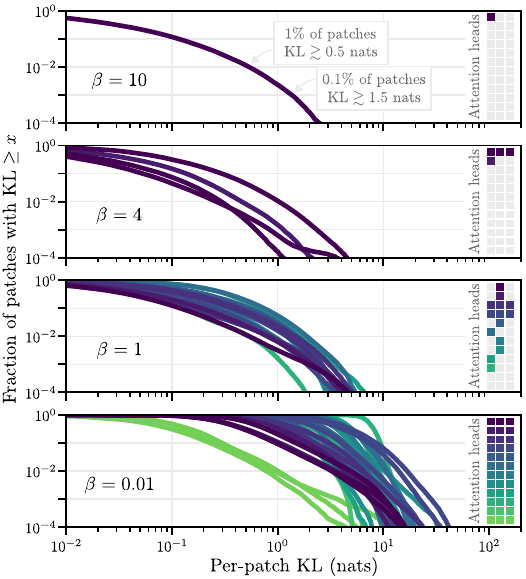}}
\caption{\textbf{Attention head firing patterns.}
Survival function 
$P(\text{KL}\ge x)$ of per-patch information across attention heads, showing that only a small fraction of patches carry significant routed information in the high-$\beta$ regime.
The grids on the right show which heads are active, starting with block 0 at the bottom.
}
\label{fig:cdfs}
\end{center}
\vskip -0.2in
\end{figure}

To quantify the correspondence between two heads with latent variables $U$ and $V$, we form a joint channel by concatenating their distributional parameters, yielding samples from $p(u,v\mid x)$.
Using the identity
\begin{equation}
I(U;V)=I(X;U)+I(X;V)-I(X;U,V),
\end{equation}
we obtain an estimate of the shared information between headwise messages.
Finally, we report a normalized mutual information,
\begin{equation}
\mathrm{NMI}(U,V)
=\frac{I(U;V)}{\sqrt{I(U;U')\,I(V;V')}},
\end{equation}
where $U'$ and $V'$ denote independent samples from the same headwise channels.
Uncertainty is propagated from the standard error of the Monte Carlo estimates of $I(X;U)$.

\subsection{Implementation details}
All experiments use the ViT-Tiny backbone from \texttt{timm} (12 blocks, embedding dimension 192, 3 attention heads per block).
We adopt this lightweight backbone to enable extensive sweeps over information budgets and detailed mechanistic analysis, while preserving the canonical transformer computation pattern.
We insert variational information bottlenecks on the attention-mediated residual writes in every block, instantiating a separate encoder-decoder bottleneck for each head in each layer (36 bottlenecks total).
This design treats each head as an independently regularized communication channel while leaving the remainder of the ViT computation unchanged.
All bottlenecks share the same architecture and latent dimensionality, differing only in their learned parameters.

For classification, we employ global average pooling over patch representations followed by the standard linear classifier head~\cite{zhai2022scaling}, ensuring that the zero-communication limit remains well-defined.

We train models on the ImageNet-100 subset introduced by \citet{cmc_hobbitlong2020}, using AdamW with learning rate $6\times10^{-4}$ and weight decay 0.05 for 1000 epochs.
Training uses a cosine learning-rate schedule with 10 epochs of warmup, and the information penalty coefficient $\beta$ is fixed over the course of training.
We apply RandAugment for standard appearance-level augmentation, but omit patch-mixing strategies such as Mixup and CutMix since these distort patch-level statistics and complicate interpretation.
Additional optimization and augmentation details are provided in Appendix~\ref{appx:implementation}.

\section{Results}
\label{results}

We present results at two complementary scales.
First, we characterize the continuum of models induced by the information cost $\beta$, demonstrating systematic changes in attention-mediated communication, predictive performance, and patch-based voting behavior.
Second, we exploit the sparsity of the low-information regime to perform fine-grained mechanistic analyses of individual attention heads.

\subsection{Properties of the spectrum of models}

The entire spectrum from independent patch processing (attention blocks are removed from the ViT) to freely flowing information as in an unmodified ViT, spans $69.2\% \pm 0.3$ to $78.2\% \pm 0.3$ top-1 validation accuracy (mean and standard deviation over three random seeds) (Fig.~\ref{fig:fig1}b).
That the attention-less baseline remains competitive
underscores that substantial visual recognition can be achieved from simple pooling of local patch evidence alone, consistent with the strong performance of local-feature-based image representations prior to deep networks \cite{lowe2004sift,sanchez2013fisher}.

Intuitively, as information communicated by attention increases, so too does classification performance.
With only around one nat of KL for every ten patches, accuracy increases by $1-2\%$.
When around one nat of KL is communicated per patch, it further increases by $\sim 2\%$.
Empirically, accuracy increases approximately linearly with $\log$ KL over much of the spectrum.
The least restricted models ($\beta = 0$) approach, though do not fully reach, the accuracy of an unmodified ViT (within $\sim$1\% top-1), suggesting a small residual effect of the IB parameterization even in the weakly constrained regime.

Without attention to pass information between patches, there can clearly be no consensus between patch representations.
How does patch independence give way to consensus as attention becomes less restricted?
We show in Fig.~\ref{fig:fig1}c characteristics of patch voting behaviors for several models close to the Pareto front.
These patch-level measures provide a complementary view of global integration: rather than only tracking top-1 accuracy, they quantify how communication induces consensus among local predictions. 
The spectrum therefore captures not just performance gains, but a qualitative transition from diverse independent patch hypotheses to coordinated global representations.

First, we characterize the variety of top-class assignments by patch, using the inverted Simpson index as an effective count of the number of classes \cite{simpson1949measurement,hill1973diversity}.
The inverted Simpson index, $1/\sum_c p_c^2$ with $p_c$ the fraction of patches with top-assigned class $c$, provides an effective number of distinct classes represented in the patch-level votes.
Second, we evaluate the total range of patch logit contributions, from min to max, as a form of opinionation of the patches comprising an image.
Independent patch processing yields a large variety of output class assignments (effective counts often exceeding 50 different classes per image) and more significant opinionation (total logit range $\gtrsim 100$) than the unmodified ViT.
Importantly, we are able to smoothly and controllably interpolate between these extremes by tuning the information cost coefficient $\beta$.

\begin{figure}[!t]
\vskip 0.2in
\begin{center}
\centerline{\includegraphics[width=\columnwidth]{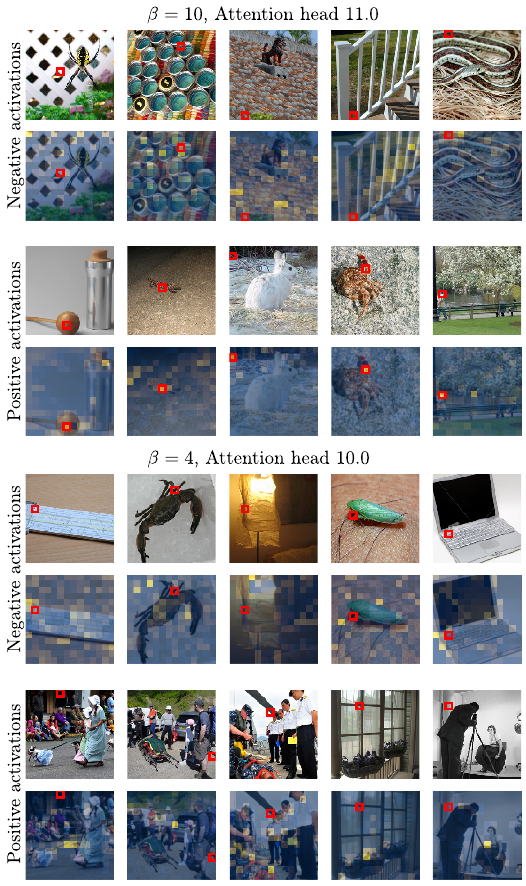}}
\caption{\textbf{Extreme activating patches.}
For the single active attention head in the $\beta=10$ model (head 11.0), and for one of the four effectual heads of the $\beta=4$ model (head 10.0), we randomly sample positive and negative activations from patches in the top 0.1\% of KL cost.
The patch is indicated by a red square in both the original image (top of each pair of rows) and the head's attention map (bottom of each pair of rows).}
\label{fig:activations}
\end{center}
\vskip -0.2in
\end{figure}

We visualize aspects of the processing of a handful of images by different models in Fig~\ref{fig:fig2} and in Appx.~\ref{appx:voting_extra}.
For the IB models, we display a heatmap of the information penalty (in terms of KL divergence) summed across all attention heads for each patch in the image.
Because each bottleneck yields a per-patch KL cost, we obtain a direct, quantitative map of where attention allocates its limited communication budget.
As is perhaps intuitive from the relatively small performance gap between independent patch processing and the full ViT (Fig.~\ref{fig:fig1}b), there is high agreement of classification probability vectors along the full spectrum for the majority of images (Appx.~\ref{appx:jsd}).
We therefore focus on images from the validation set for which ViT (unmodified) and ViT (w/o attn) vary the most in the final classification.
Specifically, we compute the Jensen-Shannon distance between the probability vectors output by the two models for a given image~\cite{lin2002divergence}, and randomly sample from the top 10\% of images.
This focuses attention on cases where global communication most strongly alters the prediction, while avoiding cherry-picking individual examples.
In Appx.~\ref{appx:voting_extra}, we show samples based on other criteria (e.g., misclassified by ViT) and additionally visualize examples of patch voting behavior.

\subsection{Update behavior of individual attention heads}
What is the contextual information required for maximal performance, that the independent patch processing cannot access?
To answer this requires interpreting the information content of the updates to the residual stream.

\begin{figure}[!t]
\vskip 0.2in
\begin{center}
\centerline{\includegraphics[width=\columnwidth]{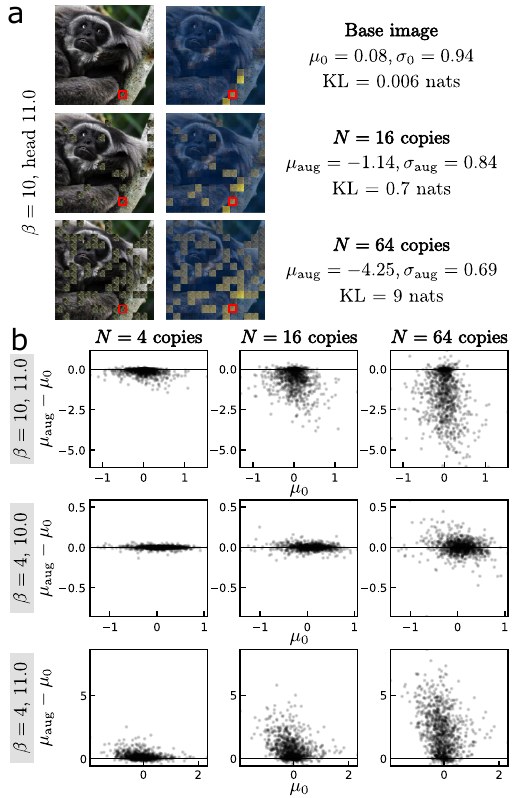}}
\caption{\textbf{Probing the repetition hypothesis.}
\textbf{(a)} Given a patch and a base image, we randomly copy the given patch to $N$ other locations in the image before passing the augmented image through the model.
For head 11.0 of the $\beta=10$ model, the repetition of the patch systematically drives the attention head's update representation to larger negative values.  
The corresponding attention maps highlight the copy locations.
\textbf{(b)} A random selection of 1024 patches in the dataset was augmented with patch copies as in \textbf{a}, at three different magnitudes and for three different attention heads (from the $\beta=10$ and $\beta=4$ models).
We display the base image's representation mean, $\mu_0$, and the signed displacement in representation space, $\mu_\text{aug}-\mu_0$, caused by the augmentation.
}
\label{fig:augs}
\end{center}
\vskip -0.2in
\end{figure}

A striking consequence of the information penalty is that many headwise channels collapse to the prior, yielding near-zero KL and effectively inactive heads.
This sparsity substantially simplifies mechanistic analysis in the low-information regime.
We show in Fig.~\ref{fig:cdfs} the survival function $\text{Pr}(\text{KL} \ge x)$ of per-patch information for all patches in the validation set and for all heads. 
We deem a head active if its KL exceeds $10^{-2}$ nats for at least one patch in the validation set.
Only one attention head is `active'' at $\beta=10$, and four heads are active for the $\beta=4$ model. 
The heads that are active spend KL sparingly: 99\% of patches receive an update costing less than 0.5 nats in the $\beta=10$ case. 
For the low-information models ($\beta=10, 4, 1$), the behavior is the same across heads, where the majority of patches get the same null update and the representation nearly mirrors the prior of the channel. 
As the information restriction loosens, $\beta=0.01$ utilizes all 36 attention heads for non-trivial updates, and the distribution of KLs spent per patch shifts significantly, to where the majority deviate from the prior.

Moreover, active heads often use only one or two latent dimensions in the low-information regime, enabling especially direct visualization of update directionality.
When the latent is effectively one-dimensional, the update reduces to a signed scalar message, allowing patches to be grouped by whether the head transmits a positive or negative signal.
In Fig.~\ref{fig:activations} we show high-update patches for the single active attention head in the $\beta=10$ model as well as the corresponding attention maps. 
Examples for other attention heads can be found in Appendix~\ref{appx:activating_patches_examples}.

The first attention head to turn on---the attention-mediated updates that are worth paying a steep information cost---appear to encode coarse statistics of patch similarity elsewhere in the image, effectively updating patches with a tag that they are part of a repeated pattern, as suggested by the negatively activating examples in Fig.~\ref{fig:activations} for $\beta=10$, head 11.0. 
Across three random seeds, all $\beta=10$ models contain an active head that appears to activate on repetition (Appx.~\ref{appx:activating_patches_examples}).
This suggests that coarse redundancy or similarity signals are among the most information-efficient messages to transmit: they are cheap to communicate, yet provide useful global context early in the spectrum.

We test this hypothesis through augmentations that change the relationship between highly activating patches and their containing image (Fig.~\ref{fig:augs}). 
For a random sample of 1024 images, and within each a randomly selected patch, we apply a stochastic augmentation that copy-pastes the selected patch multiple times elsewhere in the image. 
This manipulation increases within-image redundancy while preserving local patch content.
The effect is pronounced for the heads that seem to detect repetition: as the number of copy-pastes grows from 4 to 16 and then to 64, the mean of the latent distribution (representing the content of the attention-mediated update) continues to shift a larger amount, consistently across diverse starting patches. 
This is clear for $\beta=10$ head 11.0 and $\beta=4$ head 11.0, whereas no such systematic effect is seen for $\beta=4$ head 10.0.

As communication capacity increases, attention heads diversify in function, making single-head interpretation more nuanced.
Controlled perturbations therefore play a central role in grounding mechanistic hypotheses about the messages each head transmits.
In the case of head 11.0 of $\beta=10$, the perturbations provide convergent evidence for repetition-sensitive messaging.
We next compare the information content of the attention head updates across models with different information costs and across models with different seeds. 
Note that the probabilistic nature of the attention head updates renders prevalent point-based representation similarity measures (e.g., \cite{kornblith2019similarity,klabunde2025similarity}) problematic. 
We employ recently a proposed measure that generalizes normalized mutual information (NMI) as a comparison between hard clustering assignments to be applicable to probabilistic representation spaces \cite{nmi2024}. 
We interpret each head’s updates as inducing a soft clustering over patches conditioned on image context, and then compare the information content of the soft clustering assignments.

\begin{figure}
\vskip 0.2in
\begin{center}
\centerline{\includegraphics[width=\columnwidth]{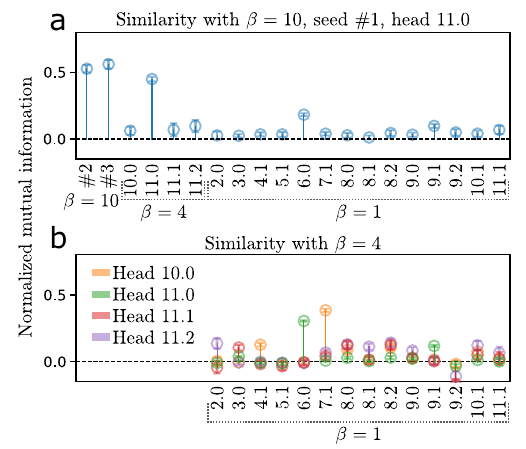}}
\caption{\textbf{Measuring the similarity of attention heads.}
\textbf{(a)} For $\beta=10$ head 11.0, we evaluate pairwise similarity with two other models trained at $\beta=10$ with different random seeds (labeled \#2 and \#3), with the four active heads in the $\beta=4$ model, and the 14 active heads in the $\beta=1$ model.
The Monte Carlo estimate for mutual information described in the main text was sampled with $2 \times 10^7$ samples; error bars show propagated uncertainty on NMI.
\textbf{(b)}
The four heads of the $\beta=4$ model were each compared to all of the heads of the $\beta=1$ model.}
\label{fig:nmi}
\end{center}
\vskip -0.2in
\end{figure}

In Fig.~\ref{fig:nmi} we show the pairwise NMI between heads of various models. 
For reference, NMI ranges from zero (no shared information) to one (identical treatment of the entire dataset). 
Across three repeats with different random seeds, the active heads of the $\beta=10$ models have high agreement, indicating a robustness of the particular tagging information seen in Fig.~\ref{fig:activations}. 
Further, the information tagged by the attention heads of $\beta=10$ is seen again, to a lesser degree, in the models trained with $\beta=4$ (head 11.0) and $\beta=1$ (head 6.0). 
The (dis)similarity of behavior under the copy-paste augmentations in Fig.~\ref{fig:augs} is seen in the corresponding NMI values: high for head 11.0 and low for head 10.0 of the $\beta=4$ model. 
When comparing the $\beta=4$ model's, four active heads to the $\beta=1$ model's fourteen active heads, there are few strong similarities worthy of deeper comparison. 

Importantly, while granular inspection of a single head is greatly aided by posterior collapse down to a single dimension, the normalized mutual information comparison is agnostic to the dimensionalities of the latent spaces \cite{nmi2024}. 
This provides a practical route for scaling mechanistic insights from highly constrained, low-dimensional channels to richer models further along the spectrum.

\section{Discussion}
\label{discussion}

Scalability remains a central challenge for mechanistic interpretability. 
Even in the most constrained regime, we interpret models with only a handful of active attention heads transmitting fractions of a bit per image. 
Yet a wide gap remains between these sparse, low-bandwidth circuits and the dense communication patterns of models approaching the performance of an unrestricted ViT (Fig.~\ref{fig:fig1}b). 
Understanding how distributed attention-based computation emerges as information constraints relax will be essential for scaling interpretability beyond highly simplified settings.

A key contribution of this work is to make attention an explicit, measurable communication channel. 
By placing variational bottlenecks directly on attention-mediated residual writes, we obtain a controlled spectrum from independent patch processing to fully expressive global attention, without otherwise modifying transformer computation (Fig.~\ref{fig:fig1}a). 
This provides a principled axis of simplification: rather than reducing model size or restricting to synthetic tasks, we restrict internal communication capacity itself.
We view this spectrum as a laboratory for tracing how attention circuits emerge as bandwidth increases.

Our results suggest that the earliest attention circuits to emerge under strong information restriction implement coarse but high-value forms of global integration. 
Across seeds, the first active heads consistently appear sensitive to patch repetition or redundancy (Figs.~\ref{fig:activations},\ref{fig:augs}), indicating that such global texture statistics may be among the most information-efficient computations available to the model. 

Several aspects of the intermediate regimes depend on modeling choices.
Our bottlenecks use Gaussian posteriors and KL regularization, and global average pooling provides a particularly clean zero-communication limit. 
Extending this framework to larger backbones, alternative aggregation structures, and richer datasets will be important for assessing the generality of the emergent behaviors we observe.

Finally, training a separate model for each information budget is computationally expensive. 
However, once trained, each point in the spectrum provides a stable object for mechanistic investigation, yielding models whose internal communication is intrinsically constrained and whose active heads are sparse and tractable. 
We view such information-restricted transformers as a useful laboratory for studying how global visual computation emerges from local processing under limited bandwidth.

\section*{Impact Statement}

This paper presents work whose goal is to advance the field of 
Machine Learning. 
There are many potential societal consequences 
of our work, none which we feel must be specifically highlighted here.

\section*{Acknowledgments}
We thank Thanh Nguyen for helpful feedback on the manuscript.

\bibliography{references}

@article{dib_pnas,
  title={Information decomposition in complex systems via machine learning},
  author={Murphy, Kieran A and Bassett, Dani S},
  journal={Proceedings of the National Academy of Sciences},
  volume={121},
  number={13},
  pages={e2312988121},
  year={2024},
  publisher={National Acad Sciences}
}

@inproceedings{
dib_iclr,
title={Interpretability with full complexity by constraining feature information},
author={Kieran A Murphy and Dani S. Bassett},
booktitle={International Conference on Learning Representations ({ICLR})},
year={2023},
url={https://openreview.net/forum?id=R_OL5mLhsv}
}

@article{
nmi2024,
title={Comparing the information content of probabilistic representation spaces},
author={Kieran A. Murphy and Sam Dillavou and Dani S. Bassett},
journal={Transactions on Machine Learning Research},
issn={2835-8856},
year={2025},
url={https://openreview.net/forum?id=adhsMqURI1},
note={}
}

@inproceedings{deepsets,
 author = {Zaheer, Manzil and Kottur, Satwik and Ravanbakhsh, Siamak and Poczos, Barnabas and Salakhutdinov, Russ R and Smola, Alexander J},
 booktitle = {Advances in Neural Information Processing Systems},
 editor = {I. Guyon and U. Von Luxburg and S. Bengio and H. Wallach and R. Fergus and S. Vishwanathan and R. Garnett},
 pages = {},
 publisher = {Curran Associates, Inc.},
 title = {Deep Sets},
 url = {https://proceedings.neurips.cc/paper/2017/file/f22e4747da1aa27e363d86d40ff442fe-Paper.pdf},
 volume = {30},
 year = {2017}
}

@article{vaswani2017,
  title={Attention is all you need},
  author={Vaswani, Ashish and Shazeer, Noam and Parmar, Niki and Uszkoreit, Jakob and Jones, Llion and Gomez, Aidan N and Kaiser, {\L}ukasz and Polosukhin, Illia},
  journal={Advances in neural information processing systems},
  volume={30},
  year={2017}
}

@inproceedings{
dosovitskiy2021,
title={An Image is Worth 16x16 Words: Transformers for Image Recognition at Scale},
author={Alexey Dosovitskiy and Lucas Beyer and Alexander Kolesnikov and Dirk Weissenborn and Xiaohua Zhai and Thomas Unterthiner and Mostafa Dehghani and Matthias Minderer and Georg Heigold and Sylvain Gelly and Jakob Uszkoreit and Neil Houlsby},
booktitle={International Conference on Learning Representations},
year={2021},
url={https://openreview.net/forum?id=YicbFdNTTy}
}

@article{saxe2019,
	doi = {10.1088/1742-5468/ab3985},
	url = {https://doi.org/10.1088\%2F1742-5468\%2Fab3985},
	year = 2019,
	month = {dec},
	publisher = {{IOP} Publishing},
	volume = {2019},
	number = {12},
	pages = {124020},
	author = {Andrew M. Saxe and Yamini Bansal and Joel Dapello and Madhu Advani and Artemy Kolchinsky and Brendan D. Tracey and David D. Cox},
	title = {On the information bottleneck theory of deep learning},
	journal = {Journal of Statistical Mechanics: Theory and Experiment},
}

@article{tishbyIB2000,
  title={The information bottleneck method},
  author={Tishby, Naftali and Pereira, Fernando C. and Bialek, William},
  journal={arXiv preprint physics/0004057},
  year={2000}
}

@inproceedings{
alemiVIB2016,
title={Deep Variational Information Bottleneck},
author={Alexander A. Alemi and Ian Fischer and Joshua V. Dillon and Kevin Murphy},
booktitle={International Conference on Learning Representations},
year={2017},
url={https://openreview.net/forum?id=HyxQzBceg}
}

@article{achillesoatto2018,
  title={Information dropout: Learning optimal representations through noisy computation},
  author={Achille, Alessandro and Soatto, Stefano},
  journal={IEEE Transactions on Pattern Analysis and Machine Intelligence},
  volume={40},
  number={12},
  pages={2897--2905},
  year={2018},
  publisher={IEEE}
}

@article{olah2020zoom,
  title={Zoom in: An introduction to circuits},
  author={Olah, Chris and Cammarata, Nick and Schubert, Ludwig and Goh, Gabriel and Petrov, Michael and Carter, Shan},
  journal={Distill},
  volume={5},
  number={3},
  pages={e00024--001},
  year={2020}
}

@article{elhage2021mathematical,
   title={A Mathematical Framework for Transformer Circuits},
   author={Elhage, Nelson and Nanda, Neel and Olsson, Catherine and Henighan, Tom and Joseph, Nicholas and Mann, Ben and Askell, Amanda and Bai, Yuntao and Chen, Anna and Conerly, Tom and DasSarma, Nova and Drain, Dawn and Ganguli, Deep and Hatfield-Dodds, Zac and Hernandez, Danny and Jones, Andy and Kernion, Jackson and Lovitt, Liane and Ndousse, Kamal and Amodei, Dario and Brown, Tom and Clark, Jack and Kaplan, Jared and McCandlish, Sam and Olah, Chris},
   year={2021},
   journal={Transformer Circuits Thread},
   note={https://transformer-circuits.pub/2021/framework/index.html}
}

@article{bricken2023monosemanticity,
   title={Towards Monosemanticity: Decomposing Language Models With Dictionary Learning},
   author={Bricken, Trenton and Templeton, Adly and Batson, Joshua and Chen, Brian and Jermyn, Adam and Conerly, Tom and Turner, Nick and Anil, Cem and Denison, Carson and Askell, Amanda and Lasenby, Robert and Wu, Yifan and Kravec, Shauna and Schiefer, Nicholas and Maxwell, Tim and Joseph, Nicholas and Hatfield-Dodds, Zac and Tamkin, Alex and Nguyen, Karina and McLean, Brayden and Burke, Josiah E and Hume, Tristan and Carter, Shan and Henighan, Tom and Olah, Christopher},
   year={2023},
   journal={Transformer Circuits Thread},
   note={https://transformer-circuits.pub/2023/monosemantic-features/index.html}
}

@inproceedings{
huben2024sparse,
title={Sparse Autoencoders Find Highly Interpretable Features in Language Models},
author={Robert Huben and Hoagy Cunningham and Logan Riggs Smith and Aidan Ewart and Lee Sharkey},
booktitle={The Twelfth International Conference on Learning Representations},
year={2024},
url={https://openreview.net/forum?id=F76bwRSLeK}
}

@article{simpson1949measurement,
  title={Measurement of diversity},
  author={Simpson, Edward H},
  journal={nature},
  volume={163},
  number={4148},
  pages={688--688},
  year={1949},
  publisher={Nature Publishing Group UK London}
}

@inproceedings{zhai2022scaling,
  title={Scaling vision transformers},
  author={Zhai, Xiaohua and Kolesnikov, Alexander and Houlsby, Neil and Beyer, Lucas},
  booktitle={Proceedings of the IEEE/CVF conference on computer vision and pattern recognition},
  pages={12104--12113},
  year={2022}
}

@inproceedings{
covert2023learning,
title={Learning to Estimate Shapley Values with Vision Transformers},
author={Ian Connick Covert and Chanwoo Kim and Su-In Lee},
booktitle={The Eleventh International Conference on Learning Representations },
year={2023},
url={https://openreview.net/forum?id=5ktFNz_pJLK}
}

@InProceedings{colors2classes,
author="Dorszewski, Teresa
and T{\v{e}}tkov{\'a}, Lenka
and Jenssen, Robert
and Hansen, Lars Kai
and Wickstr{\o}m, Kristoffer Knutsen",
editor="Guidotti, Riccardo
and Schmid, Ute
and Longo, Luca",
title="From Colors to Classes: Emergence of Concepts in Vision Transformers",
booktitle="Explainable Artificial Intelligence",
year="2026",
publisher="Springer Nature Switzerland",
pages="28--47"
}

@inproceedings{
park2023what,
title={What Do Self-Supervised Vision Transformers Learn?},
author={Namuk Park and Wonjae Kim and Byeongho Heo and Taekyung Kim and Sangdoo Yun},
booktitle={The Eleventh International Conference on Learning Representations },
year={2023},
url={https://openreview.net/forum?id=azCKuYyS74}
}

@inproceedings{koh2020cbm,
  title={Concept bottleneck models},
  author={Koh, Pang Wei and Nguyen, Thao and Tang, Yew Siang and Mussmann, Stephen and Pierson, Emma and Kim, Been and Liang, Percy},
  booktitle={International conference on machine learning},
  pages={5338--5348},
  year={2020},
  organization={PMLR}
}

@article{pan2021ia,
  title={IA-RED\textsuperscript{2}: Interpretability-aware redundancy reduction for vision transformers},
  author={Pan, Bowen and Panda, Rameswar and Jiang, Yifan and Wang, Zhangyang and Feris, Rogerio and Oliva, Aude},
  journal={Advances in neural information processing systems},
  volume={34},
  pages={24898--24911},
  year={2021}
}

@article{hu2024survey,
  title={A survey on information bottleneck},
  author={Hu, Shizhe and Lou, Zhengzheng and Yan, Xiaoqiang and Ye, Yangdong},
  journal={IEEE Transactions on Pattern Analysis and Machine Intelligence},
  volume={46},
  number={8},
  pages={5325--5344},
  year={2024},
  publisher={IEEE}
}

@article{chen2019looks,
  title={This looks like that: deep learning for interpretable image recognition},
  author={Chen, Chaofan and Li, Oscar and Tao, Daniel and Barnett, Alina and Rudin, Cynthia and Su, Jonathan K},
  journal={Advances in neural information processing systems},
  volume={32},
  year={2019}
}

@article{lowe2004sift,
  title={Distinctive image features from scale-invariant keypoints},
  author={Lowe, David G},
  journal={International journal of computer vision},
  volume={60},
  number={2},
  pages={91--110},
  year={2004},
  publisher={Springer}
}

@article{sanchez2013fisher,
  title={Image classification with the fisher vector: Theory and practice},
  author={S{\'a}nchez, Jorge and Perronnin, Florent and Mensink, Thomas and Verbeek, Jakob},
  journal={International journal of computer vision},
  volume={105},
  number={3},
  pages={222--245},
  year={2013},
  publisher={Springer}
}

@article{lin2002divergence,
  title={Divergence measures based on the Shannon entropy},
  author={Lin, Jianhua},
  journal={IEEE Transactions on Information theory},
  volume={37},
  number={1},
  pages={145--151},
  year={2002},
  publisher={IEEE}
}

@article{hill1973diversity,
  title={Diversity and evenness: a unifying notation and its consequences},
  author={Hill, Mark O},
  journal={Ecology},
  volume={54},
  number={2},
  pages={427--432},
  year={1973},
  publisher={Wiley Online Library}
}

@article{rudin2019stop,
  title={Stop explaining black box machine learning models for high stakes decisions and use interpretable models instead},
  author={Rudin, Cynthia},
  journal={Nature Machine Intelligence},
  volume={1},
  number={5},
  pages={206--215},
  year={2019},
  publisher={Nature Publishing Group}
}

@inproceedings{kornblith2019similarity,
  title={Similarity of neural network representations revisited},
  author={Kornblith, Simon and Norouzi, Mohammad and Lee, Honglak and Hinton, Geoffrey},
  booktitle={International conference on machine learning},
  pages={3519--3529},
  year={2019},
  organization={PMLR}
}

@inproceedings{chefer2021deeptaylor,
  title={Transformer interpretability beyond attention visualization},
  author={Chefer, Hila and Gur, Shir and Wolf, Lior},
  booktitle={Proceedings of the IEEE/CVF conference on computer vision and pattern recognition},
  pages={782--791},
  year={2021}
}

@InProceedings{lrp_transformers_2024,
  title = {{A}ttn{LRP}: Attention-Aware Layer-Wise Relevance Propagation for Transformers},
  author = {Achtibat, Reduan and Hatefi, Sayed Mohammad Vakilzadeh and Dreyer, Maximilian and Jain, Aakriti and Wiegand, Thomas and Lapuschkin, Sebastian and Samek, Wojciech},
  booktitle = {Proceedings of the 41st International Conference on Machine Learning},
  pages = {135--168},
  year = {2024},
  editor = {Salakhutdinov, Ruslan and Kolter, Zico and Heller, Katherine and Weller, Adrian and Oliver, Nuria and Scarlett, Jonathan and Berkenkamp, Felix},
  volume = {235},
  series = {Proceedings of Machine Learning Research},
  month = {21--27 Jul},
  publisher = {PMLR}
}

@article{rao2021dynamicvit,
  title={Dynamicvit: Efficient vision transformers with dynamic token sparsification},
  author={Rao, Yongming and Zhao, Wenliang and Liu, Benlin and Lu, Jiwen and Zhou, Jie and Hsieh, Cho-Jui},
  journal={Advances in neural information processing systems},
  volume={34},
  pages={13937--13949},
  year={2021}
}

@article{liang2022evit,
  title={Not all patches are what you need: Expediting vision transformers via token reorganizations},
  author={Liang, Youwei and Ge, Chongjian and Tong, Zhan and Song, Yibing and Wang, Jue and Xie, Pengtao},
  journal={arXiv preprint arXiv:2202.07800},
  year={2022}
}

@article{michel2019pruning,
  title={Are sixteen heads really better than one?},
  author={Michel, Paul and Levy, Omer and Neubig, Graham},
  journal={Advances in neural information processing systems},
  volume={32},
  year={2019}
}

@article{voita2019analyzing,
  title={Analyzing multi-head self-attention: Specialized heads do the heavy lifting, the rest can be pruned},
  author={Voita, Elena and Talbot, David and Moiseev, Fedor and Sennrich, Rico and Titov, Ivan},
  journal={arXiv preprint arXiv:1905.09418},
  year={2019}
}

@article{raghu2021vision,
  title={Do vision transformers see like convolutional neural networks?},
  author={Raghu, Maithra and Unterthiner, Thomas and Kornblith, Simon and Zhang, Chiyuan and Dosovitskiy, Alexey},
  journal={Advances in neural information processing systems},
  volume={34},
  pages={12116--12128},
  year={2021}
}

@inproceedings{caron2021emerging,
  title={Emerging properties in self-supervised vision transformers},
  author={Caron, Mathilde and Touvron, Hugo and Misra, Ishan and J{\'e}gou, Herv{\'e} and Mairal, Julien and Bojanowski, Piotr and Joulin, Armand},
  booktitle={Proceedings of the IEEE/CVF international conference on computer vision},
  pages={9650--9660},
  year={2021}
}

@article{klabunde2025similarity,
  title={Similarity of neural network models: A survey of functional and representational measures},
  author={Klabunde, Max and Schumacher, Tobias and Strohmaier, Markus and Lemmerich, Florian},
  journal={ACM Computing Surveys},
  volume={57},
  number={9},
  pages={1--52},
  year={2025},
  publisher={ACM New York, NY}
}

@article{samek2021xai,
  title={Explaining deep neural networks and beyond: A review of methods and applications},
  author={Samek, Wojciech and Montavon, Gr{\'e}goire and Lapuschkin, Sebastian and Anders, Christopher J and M{\"u}ller, Klaus-Robert},
  journal={Proceedings of the IEEE},
  volume={109},
  number={3},
  pages={247--278},
  year={2021},
  publisher={IEEE}
}

@article{
dutta2024how,
title={How to think step-by-step: A mechanistic understanding of chain-of-thought reasoning},
author={Subhabrata Dutta and Joykirat Singh and Soumen Chakrabarti and Tanmoy Chakraborty},
journal={Transactions on Machine Learning Research},
issn={2835-8856},
year={2024},
url={https://openreview.net/forum?id=uHLDkQVtyC},
note={}
}

@inproceedings{
akyurek2023what,
title={What learning algorithm is in-context learning? Investigations with linear models},
author={Ekin Aky{\"u}rek and Dale Schuurmans and Jacob Andreas and Tengyu Ma and Denny Zhou},
booktitle={The Eleventh International Conference on Learning Representations },
year={2023},
url={https://openreview.net/forum?id=0g0X4H8yN4I}
}

@inproceedings{
nanda2023grok,
title={Progress measures for grokking via mechanistic interpretability},
author={Neel Nanda and Lawrence Chan and Tom Lieberum and Jess Smith and Jacob Steinhardt},
booktitle={The Eleventh International Conference on Learning Representations },
year={2023},
url={https://openreview.net/forum?id=9XFSbDPmdW}
}

@inproceedings{
wang2023IOI,
title={Interpretability in the Wild: a Circuit for Indirect Object Identification in {GPT}-2 Small},
author={Kevin Ro Wang and Alexandre Variengien and Arthur Conmy and Buck Shlegeris and Jacob Steinhardt},
booktitle={The Eleventh International Conference on Learning Representations },
year={2023},
url={https://openreview.net/forum?id=NpsVSN6o4ul}
}

@article{stevens2025SAEViT,
  title={Sparse autoencoders for scientifically rigorous interpretation of vision models},
  author={Stevens, Samuel and Chao, Wei-Lun and Berger-Wolf, Tanya and Su, Yu},
  journal={arXiv preprint arXiv:2502.06755},
  year={2025}
}

@inproceedings{
Schulz2020IBA,
title={Restricting the Flow: Information Bottlenecks for Attribution},
author={Karl Schulz and Leon Sixt and Federico Tombari and Tim Landgraf},
booktitle={International Conference on Learning Representations},
year={2020},
url={https://openreview.net/forum?id=S1xWh1rYwB}
}

@inproceedings{jiang2020IBAEMNLP,
    title = "{I}nserting {I}nformation {B}ottlenecks for {A}ttribution in {T}ransformers",
    author = "Jiang, Zhiying  and
      Tang, Raphael  and
      Xin, Ji  and
      Lin, Jimmy",
    booktitle = "Findings of the Association for Computational Linguistics: EMNLP 2020",
    month = nov,
    year = "2020",
    address = "Online",
    publisher = "Association for Computational Linguistics",
    url = "https://www.aclweb.org/anthology/2020.findings-emnlp.343",
    pages = "3850--3857",
}

@INPROCEEDINGS{Hong2025CoIBA,
  author={Hong, Jung-Ho and Kim, Ho-Joong and Jeon, Kyu-Sung and Lee, Seong-Whan},
  booktitle={2025 IEEE/CVF Conference on Computer Vision and Pattern Recognition (CVPR)}, 
  title={Comprehensive Information Bottleneck for Unveiling Universal Attribution to Interpret Vision Transformers}, 
  year={2025},
  volume={},
  number={},
  pages={25166-25175},
  doi={10.1109/CVPR52734.2025.02343},
}

@InProceedings{cmc_hobbitlong2020,
author="Tian, Yonglong
and Krishnan, Dilip
and Isola, Phillip",
editor="Vedaldi, Andrea
and Bischof, Horst
and Brox, Thomas
and Frahm, Jan-Michael",
title="Contrastive Multiview Coding",
booktitle="Computer Vision -- ECCV 2020",
year="2020",
publisher="Springer International Publishing",
address="Cham",
pages="776--794",
isbn="978-3-030-58621-8"
}

@inproceedings{randaugment,
 author = {Cubuk, Ekin Dogus and Zoph, Barret and Shlens, Jon and Le, Quoc},
 booktitle = {Advances in Neural Information Processing Systems},
 editor = {H. Larochelle and M. Ranzato and R. Hadsell and M.F. Balcan and H. Lin},
 pages = {18613--18624},
 publisher = {Curran Associates, Inc.},
 title = {RandAugment: Practical Automated Data Augmentation with a Reduced Search Space},
 url = {https://proceedings.neurips.cc/paper_files/paper/2020/file/d85b63ef0ccb114d0a3bb7b7d808028f-Paper.pdf},
 volume = {33},
 year = {2020}
}
\bibliographystyle{icml2026}

\newpage
\appendix
\onecolumn
\section{Implementation details}
\label{appx:implementation}

Accompanying code may be found on \href{https://github.com/murphyka/vit_ib}{github}.

All experiments are based on Vision Transformer (ViT) models implemented using the \texttt{timm} library. 
Unless otherwise stated, we use standard ViT architectures (e.g., patch size 16, image resolution 224) with modifications described below.

To study information-restricted attention, we wrap the attention module to bottleneck the per-head residual updates. 
For every head, the message passes through a variational information bottleneck (IB) module before being written to the residual stream. 
The IB is applied independently per attention head and per token, and returns both the transformed messages and a KL-divergence term used for regularization during training.
For classification, we use global average pooling over the token representations and then apply the final projection to class logits.
The remainder of the ViT architecture (patch embedding, positional embeddings, MLP blocks, normalization layers) follows the standard ViT design.

\textbf{Optimization.} 
Models are trained using the AdamW optimizer with a base learning rate of $6\times10^{-4}$ and a weight decay of 0.05, for 1000 epochs.
Bias parameters, parameters with dimensionality one (e.g., LayerNorm scale parameters), and all parameters in the IB encoder and decoder are excluded from weight decay. 
All other parameters, including patch embeddings and positional embeddings, are subject to weight decay.
Automatic mixed precision (AMP) is used with gradient scaling.

Training uses a cosine learning-rate schedule with linear warmup over the first 10 epochs. 
The coefficient $\beta$ is fixed over the course of training.

\textbf{Data and Augmentation.}
Experiments are conducted on the 100-class subset of Imagenet used in \citet{cmc_hobbitlong2020} and specified in the accompanying \href{https://github.com/HobbitLong/CMC/blob/master/imagenet100.txt}{github repository}. 
Training images are augmented using random resized cropping, horizontal flipping, and RandAugment~\cite{randaugment}. 
We use a random resized crop with scale range [0.08, 1.0] and RandAugment with two augmentation operations of magnitude 9. 
Validation images are resized and center-cropped without stochastic augmentation. 
All images are normalized using standard ImageNet mean and standard deviation values.

\section{Similarity of classification outputs between models in the spectrum}
\label{appx:jsd}
\begin{figure}[ht]
\vskip 0.2in
\begin{center}
\centerline{\includegraphics[width=0.6\columnwidth]{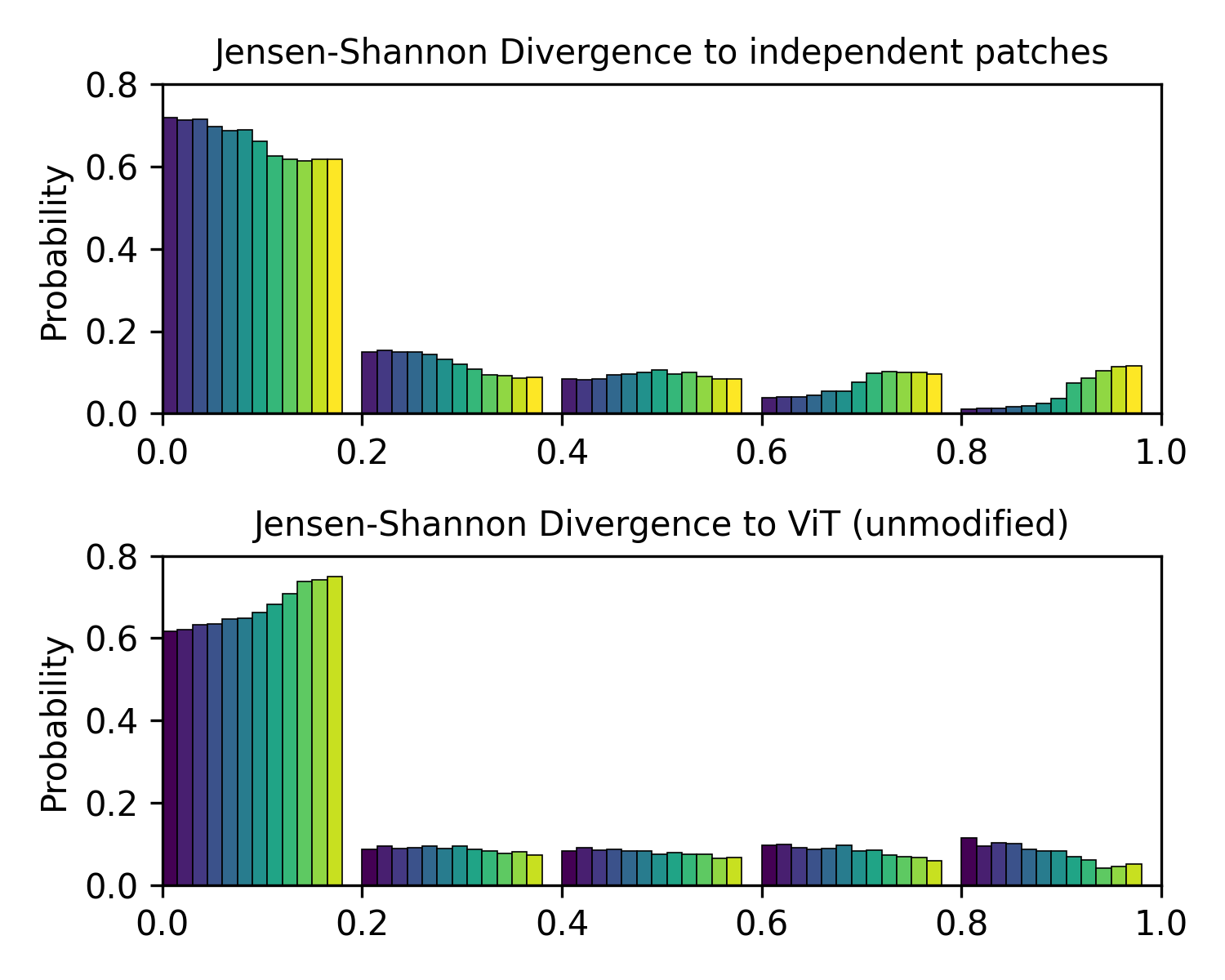}}
\caption{\textbf{Jensen-Shannon divergence between classification outputs across the spectrum.}
For all images in the validation set, we measure the similarity of model classification outputs across the spectrum to that of the independent patch processing model (ViT w/o attention, dark blue, top row) and to that of the unmodified ViT (yellow, bottom row).}
\label{fig:appx_jsd}
\end{center}
\vskip -0.2in
\end{figure}

We compared the similarity of classification probability vectors for models across the spectrum, using the Jensen-Shannon divergence.
The divergence values, grouped into five bins, are shown in Fig.~\ref{fig:appx_jsd}.

\section{Additional comparisons between model updates and patch voting}
\label{appx:voting_extra}
\begin{figure}[ht]
\begin{center}
\centerline{\includegraphics[width=\columnwidth]{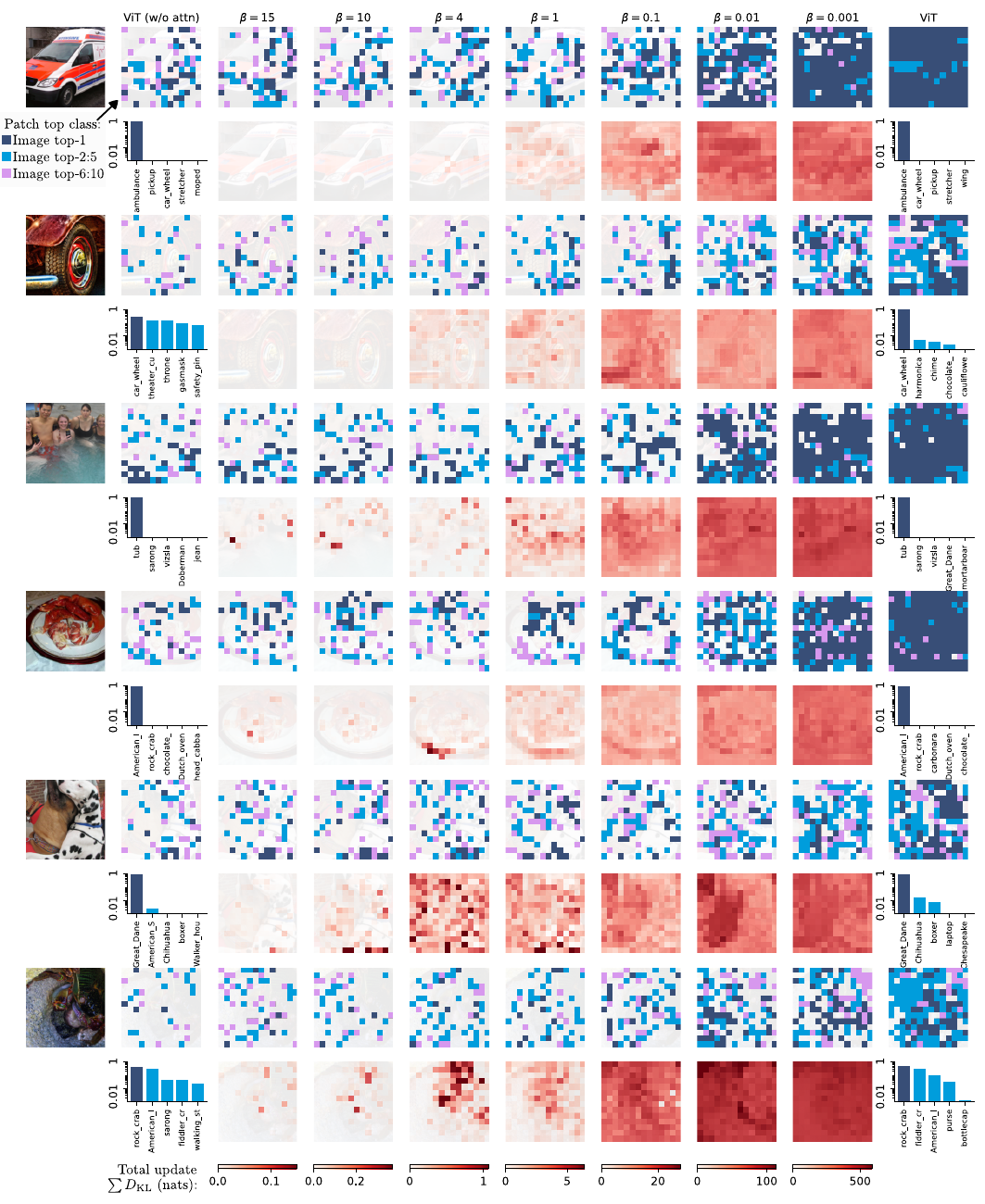}}
\caption{\textbf{Patch voting and attention head update behavior for a random selection of images correctly classified by ViT (seed \#1).}
Patches are colored according to the particular model's class assignments: dark blue for the patches whose top class matches that of the full image prediction top class, light blue for ranks 2 through 5, and pink for ranks 6 through 10.
}
\label{fig:appx_grp0}
\end{center}
\vskip -0.2in
\end{figure}

\begin{figure}[ht]
\begin{center}
\centerline{\includegraphics[width=\columnwidth]{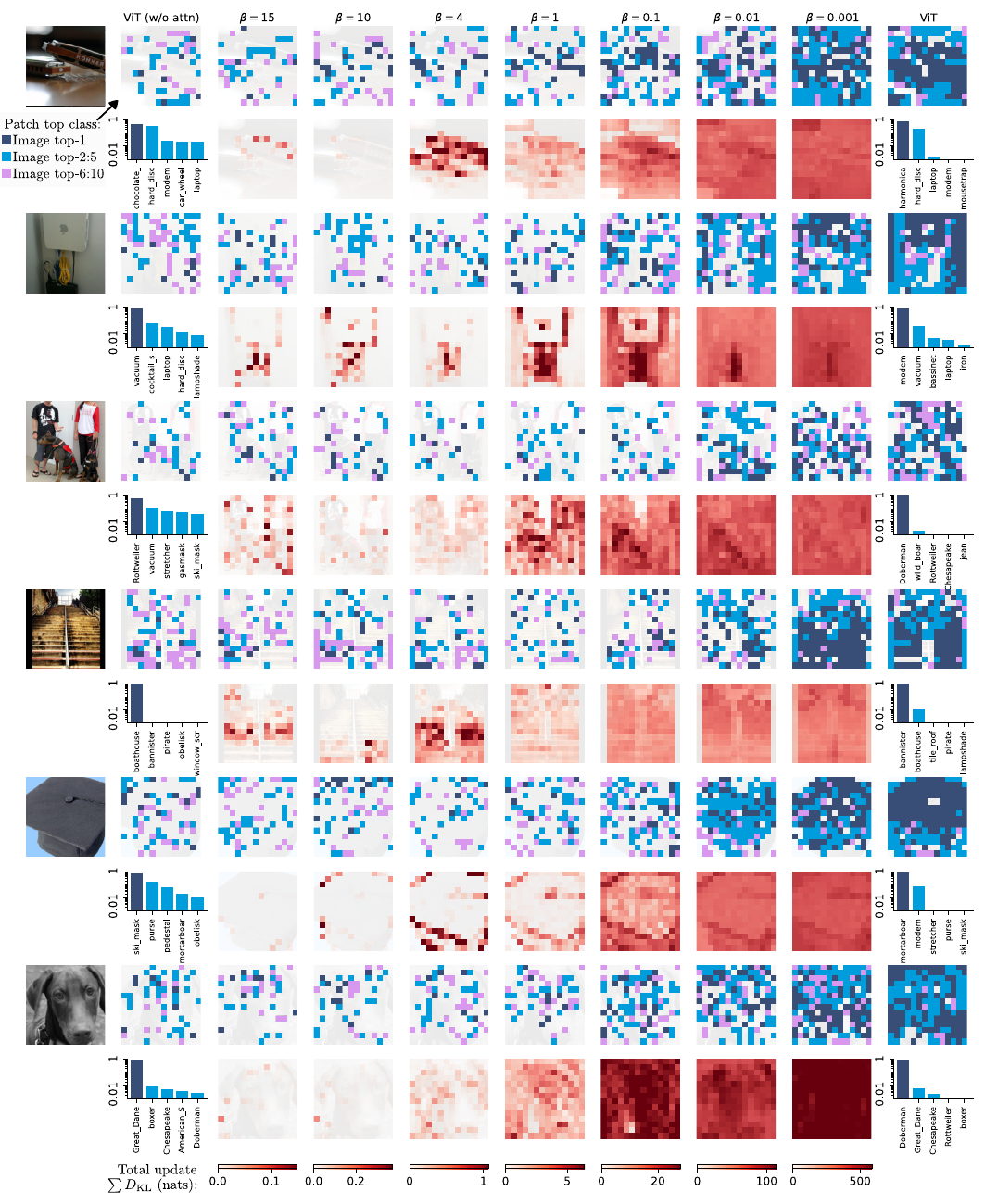}}
\caption{\textbf{Patch voting and attention head update behavior for a random selection of images correctly classified by ViT (seed \#1) \texttt{AND} in the top 10\% of images in terms of JSD between the ViT without attention and the unmodified ViT.}}
\label{fig:appx_grp1}
\end{center}
\vskip -0.2in
\end{figure}

\begin{figure}[ht]
\begin{center}
\centerline{\includegraphics[width=\columnwidth]{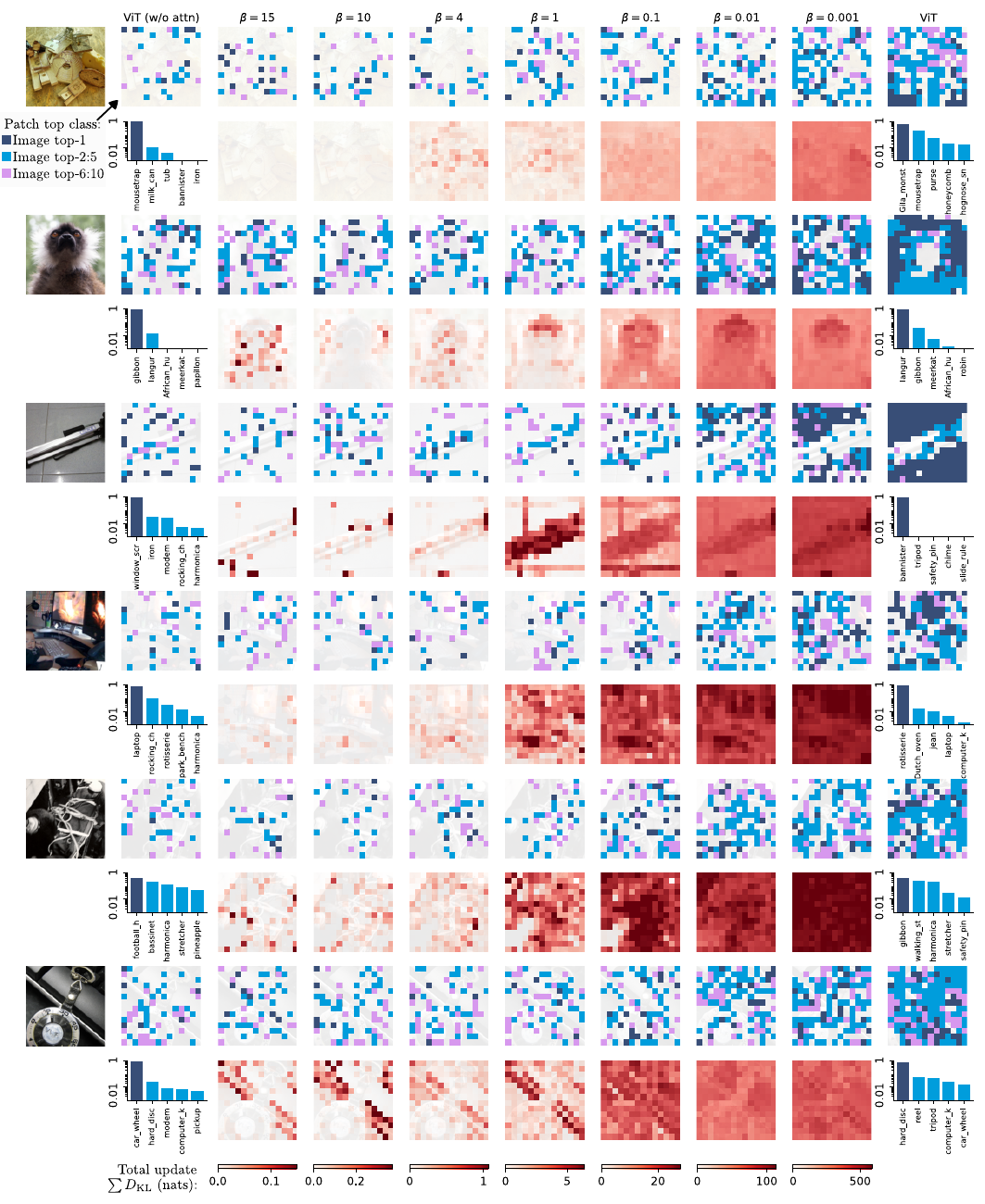}}
\caption{\textbf{Patch voting and attention head update behavior for a random selection of images incorrectly classified by ViT (seed \#1)}}
\label{fig:appx_grp2}
\end{center}
\vskip -0.2in
\end{figure}

In Figs.~\ref{fig:appx_grp0}, \ref{fig:appx_grp1}, and \ref{fig:appx_grp2}, we show how different models along the spectrum ``vote'' for different images, and for the information restricted models, how the KL budget is spent.

\section{Additional attention head highly activating patches}
\begin{figure}[ht]
\begin{center}
\centerline{\includegraphics[width=0.95\columnwidth]{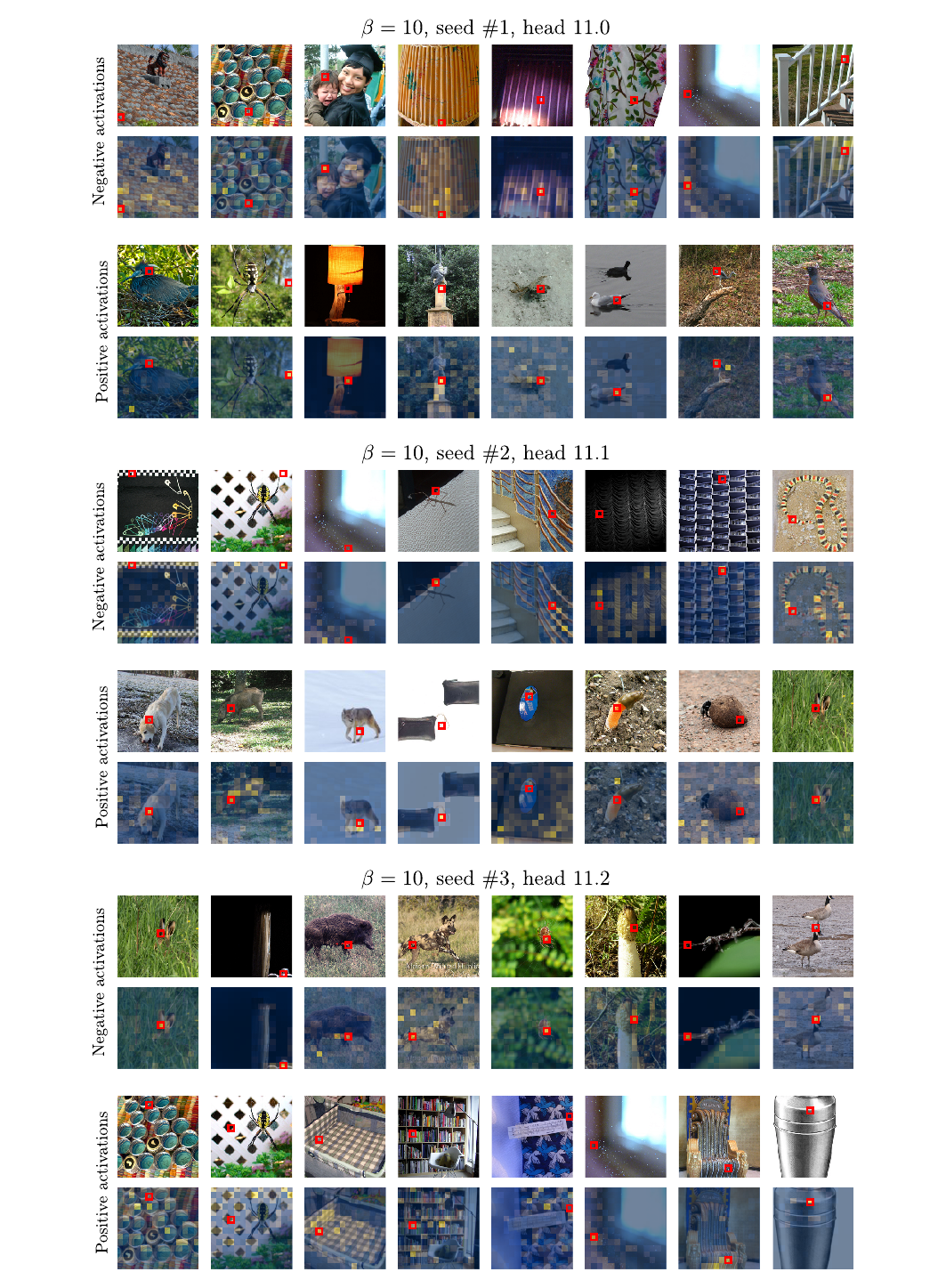}}
\caption{\textbf{Additional top activation examples for heads with one active latent dimension: different $\beta=10$ seeds.}}
\label{fig:appx_acts_beta10}
\end{center}
\vskip -0.2in
\end{figure}

\begin{figure}[ht]
\begin{center}
\centerline{\includegraphics[width=0.95\columnwidth]{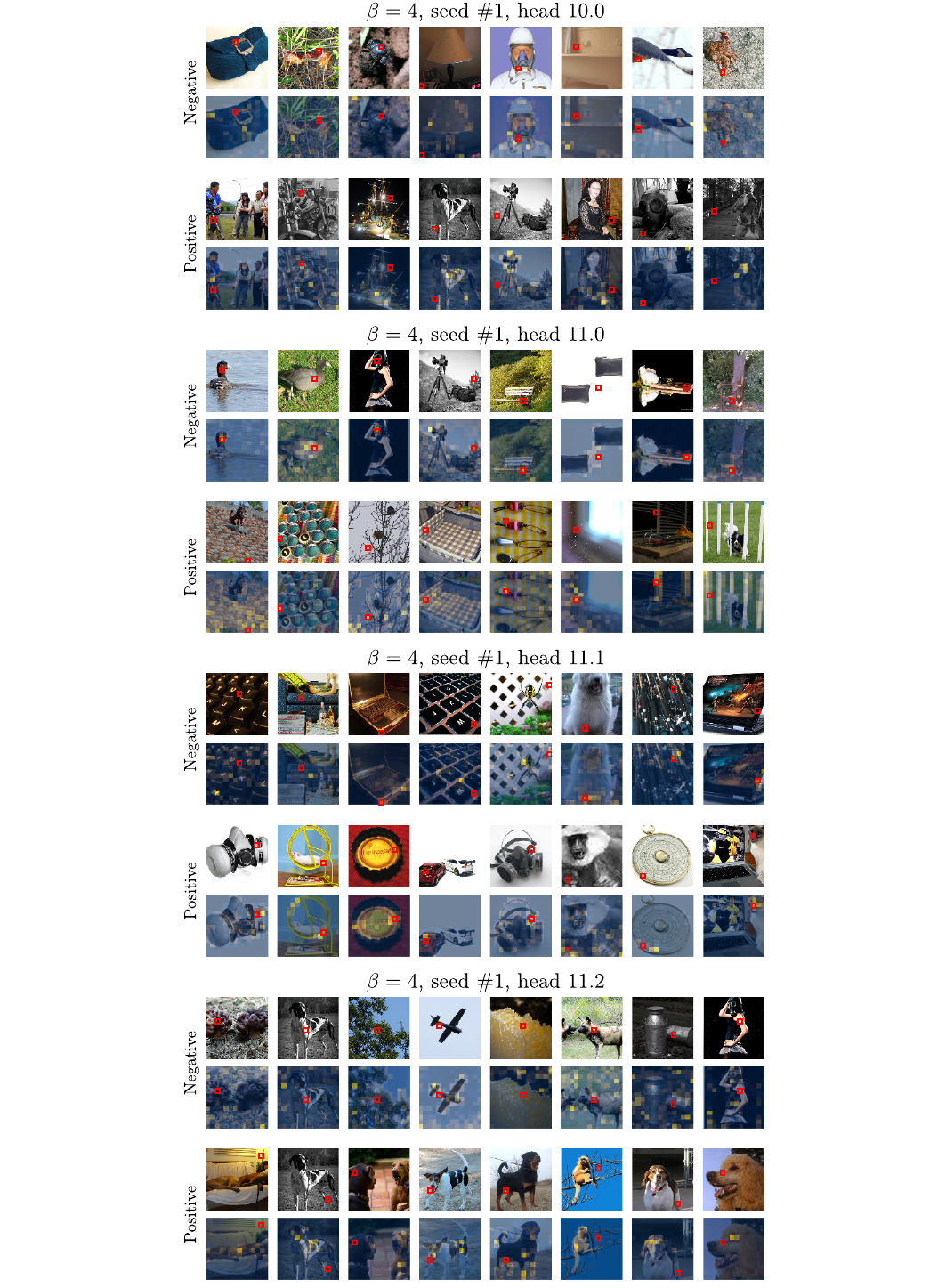}}
\caption{\textbf{Additional top activation examples for heads with one active latent dimension: different heads in the $\beta=4$ model.}}
\label{fig:appx_acts_beta4}
\end{center}
\vskip -0.2in
\end{figure}

In Figs.~\ref{fig:appx_acts_beta10}\&\ref{fig:appx_acts_beta4}, we show top activating patches of both polarities for other seeds at $\beta=10$ and for the four active heads of the $\beta=4$ model.
\label{appx:activating_patches_examples}

\end{document}